\pdfoutput=1

\documentclass[11pt]{article}

\usepackage[]{ACL2023}

\usepackage{times}
\usepackage{latexsym}
\usepackage{graphicx}
\usepackage[utf8]{inputenc} 
\usepackage[T1]{fontenc}    
\usepackage{hyperref}       
\usepackage{url}            
\usepackage{booktabs}       
\usepackage{amsfonts}       
\usepackage{nicefrac}       
\usepackage{microtype}      
\usepackage{xcolor}         
\usepackage{graphicx}
\usepackage{amsmath}
\usepackage{amsfonts}
\usepackage{multirow}
\usepackage{amsfonts}
\usepackage{tikz}
\usetikzlibrary{calc}
\usepackage{tikz-qtree}
\usetikzlibrary{trees}
\usepackage[edges]{forest}
\usetikzlibrary{shadows,arrows.meta}
\usepackage{makecell}
\usepackage{color}
\usepackage{CJKutf8}
\usepackage{rotating}

\usepackage{siunitx}
\usepackage{colortbl}
\usepackage{makecell}
\usepackage[ruled,linesnumbered]{algorithm2e}
\usepackage{algpseudocode} 
\usepackage{subcaption}
\usepackage{bbm}
\definecolor{maroon}{cmyk}{0,0.87,0.68,0.32}
\usepackage{colortbl}

\usepackage[T1]{fontenc}

\usepackage[utf8]{inputenc}

\usepackage{microtype}

\usepackage{inconsolata}

\title{UNIMO-G: Unified Image Generation through Multimodal Conditional Diffusion}

\author{ Wei Li$^*$, Xue Xu$^*$, Jiachen Liu, Xinyan Xiao \\
  Baidu Inc., Beijing, China \\
  \texttt{\{liwei85.2023\}@gmail.com}\\
  \texttt{\{xuxue,xiaoxinyan,liujiachen\}@baidu.com}\\
  }

\begin{document}
\maketitle
\def\thefootnote{*}\footnotetext{These authors contributed equally to this work.}
\def\thefootnote{}\footnotetext{\url{https://unimo-ptm.github.io/}}
\begin{abstract}
Existing text-to-image diffusion models primarily generate images from text prompts. However, the inherent conciseness of textual descriptions poses challenges in faithfully synthesizing images with intricate details, such as specific entities or scenes. This paper presents \textbf{UNIMO-G}, a simple multimodal conditional diffusion framework that operates on multimodal prompts with interleaved textual and visual inputs, which demonstrates a unified ability for both text-driven and subject-driven image generation. UNIMO-G comprises two core components: a Multimodal Large Language Model (MLLM) for encoding multimodal prompts, and a conditional denoising diffusion network for generating images based on the encoded multimodal input. We leverage a two-stage training strategy to effectively train the framework: firstly pre-training on large-scale text-image pairs to develop conditional image generation capabilities, and then instruction tuning with multimodal prompts to achieve unified image generation proficiency. A well-designed data processing pipeline involving language grounding and image segmentation is employed to construct multi-modal prompts. UNIMO-G excels in both text-to-image generation and zero-shot subject-driven synthesis, and is notably effective in generating high-fidelity images from complex multimodal prompts involving multiple image entities.
\end{abstract}

\section{Introduction}

\begin{figure*}[!ht]
    \centering
    \includegraphics[width=0.9\linewidth]{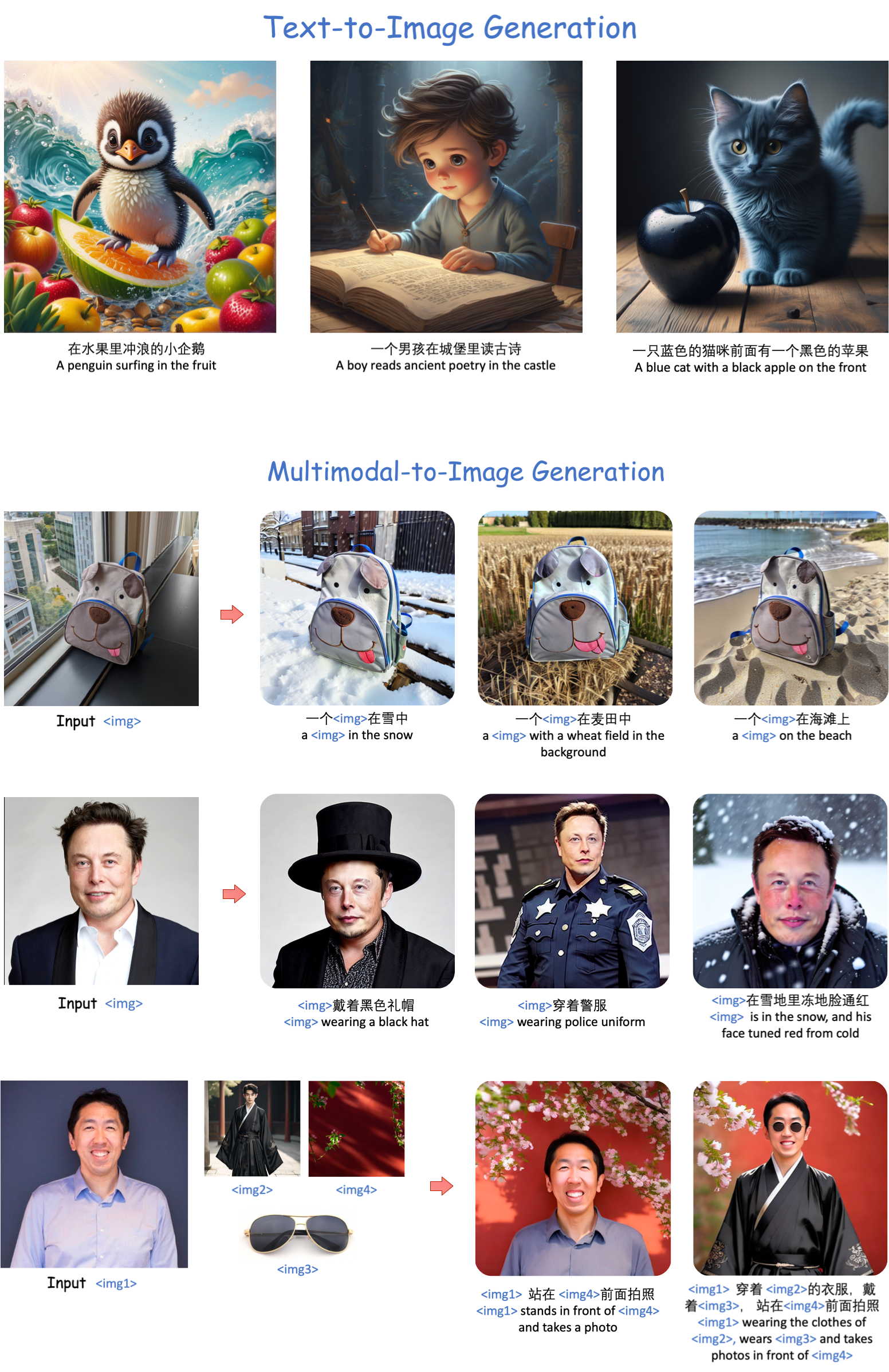}
    \caption{Examples of UNIMO-G for both text-driven and zero-shot subject-driven generation. UNIMO-G can perceive free-form interleaved visual-language inputs and faithfully generate images. Particularly, it can generate images from multi-modal prompts with multiple image entities.}
    \label{fig:showcase}
\end{figure*}

Recent advancements in text-to-image (T2I) diffusion models have yielded impressive results in the generation of high-fidelity images from textual descriptions. Various methods, including DALL-Es~\cite{ramesh2022hierarchical,betker2023improving}, Imagen~\cite{saharia2022photorealistic}, Stable Diffusion~\cite{rombach2022high}, and MM-DiT~\cite{esser2024scaling}, have been successful in producing photorealistic and contextually relevant images based on textual prompts. Nevertheless, a fundamental challenge persists due to the inherent brevity of textual descriptions, particularly when intricate details, specific entities, or nuanced scenes are involved. Thus, faithfully generating images from general vision-language (VL) inputs is essential to improve the controllability of image generation. 

Numerous studies have explored VL-to-image generation techniques. Methods such as DreamBooth~\cite{ruiz2023dreambooth}, Imagic~\cite{kawar2023imagic}, SuTI~\cite{chen2023subject} and BLIP-Diffusion~\cite{li2023blip} emphasize subject-driven generation, where they use both subject images and textual descriptions as inputs to recontextualize the subject in a newly described setting. They either fine-tune specific models for a given subject or employ pre-trained subject representations. However, their specific training design and input templates hinder their scalability, especially in complex scenarios with multiple entities. Additionally, studies like FastComposer~\cite{xiao2023fastcomposer} and Subject-Diffusion~\cite{ma2023subject} focus on multiple-entity image generation, integrating image embeddings from image encoders with the standard text conditioning in pre-trained diffusion models. Nevertheless, these approaches lack the capacity to efficiently process generalized vision-language inputs that comprise a mix of textual and visual information in free forms.

In this paper, we propose \textbf{UNIMO-G}, a simple multimodal conditional diffusion framework that operates on multimodal prompts comprising free-form interleaved vision-language inputs. Unlike traditional text-only prompts, multimodal prompts encompass various combinations of image entities and textual elements, as demonstrated in Figure~\ref{fig:showcase}. UNIMO-G is designed to faithfully reproduce all image entities, render textual content, and follow the instructions in multimodal prompts. Specifically, we leverage the perception capabilities of Multimodal Large Language Models (MLLMs) to encode multimodal prompts into a unified vision-language semantic space. Subsequently, a conditional diffusion network generates images from these encoded representations. 

To train UNIMO-G efficiently, we implement a two-phase strategy. Initially, the model undergoes pre-training on a large-scale dataset of text-image pairs, enhancing its proficiency in conditional image generation. This is followed by a phase of instruction tuning with multimodal prompts, learns to generate images that align with the detailed specifications provided in these prompts. A carefully designed data processing pipeline, incorporating language grounding and image segmentation, is employed to construct these multimodal prompts. This approach enables UNIMO-G to harness rich features from the MLLM encoder to generate images faithfully reproducing the contents across various contexts. 

UNIMO-G exhibits a comprehensive capability for controllable image generation, excelling not only in text-to-image synthesis but also in zero-shot subject-driven generation. It adeptly produces high-fidelity images from multimodal prompts, even those containing multiple image entities. To assess its performance, we conducted evaluations in both text-to-image and subject-driven generation contexts using the MS-COCO~\cite{lin2014microsoft} and DreamBench~\cite{ruiz2023dreambooth} datasets, respectively. The results consistently highlight UNIMO-G's superior performance in these scenarios. Additionally, recognizing DreamBench's focus on single-subject generation, we introduce MultiBench, a new benchmark featuring images with multiple entities. The evaluation on MultiBench confirms UNIMO-G's effectiveness in zero-shot multi-entity subject-driven generation.

In summary, our contributions in this work can be summarized as follows:
\begin{itemize}
    \item We propose a simple multi-modal conditional diffusion framework that significantly enhances the controllability of image generation by supporting multimodal prompts with interleaved images and text input.
    \item We introduce an effective two-stage training strategy, empowering zero-shot multi-entity subject-driven generation through multi-modal instruction tuning.
    \item UNIMO-G outperforms existing VL-to-image models in both single and multi-entity subject-driven generation tasks, especially on the capabilities of multimodal instruction following.
\end{itemize}

\section{Method}
\label{method}
\begin{figure}
    \centering
    \includegraphics[width=0.95\linewidth]{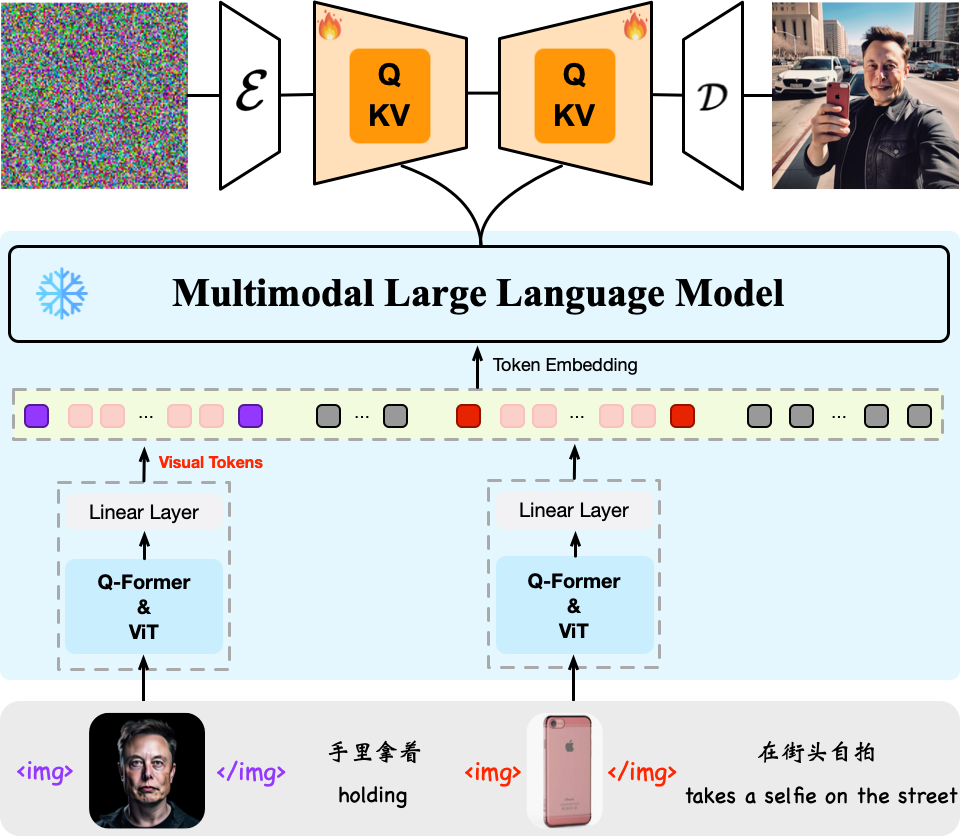}
    \caption{UNIMO-G consists of an MLLM for multimodal perception, and a conditional denoising UNet for image generation. It accepts multimodal prompts with interleaved images and texts, and generates images consistent with the image entities. Orange denotes the trainable modules; Blue denotes the frozen ones.}
    \label{fig:architecture}
\end{figure}

The architecture of UNIMO-G, as depicted in Figure~\ref{fig:architecture}, primarily comprises two key components: a Multimodal Large Language Model (MLLM) responsible for encoding multimodal prompts and a conditional denoising diffusion network for image generation based on the encoded representations. 
In our study, we employed a pre-trained Chinese MLLM, structurally similar to MiniGPT-4~\cite{zhu2023minigpt}. 
It consists of a vision encoder with a pretrained ViT and Q-Former, a single linear projection layer, and a Transformer-based LLM, underwent pre-training on a vast dataset comprising billions of image-text pairs. Details can refer to Appendix~\ref{sec:mllm}. This extensive training process equips the model with a robust capability to process and interpret complex multimodal data.

The training of UNIMO-G is conducted in a two-stage process:
\begin{itemize}
    \item \textbf{Text-to-Image Pre-training}: We pre-train the conditional denoising diffusion network from scratch on large-scale Chinese text-image pairs. We employ the same U-Net network architecture in~\citet{rombach2022high} and condition it on the text using a cross-attention mechanism.
    \item \textbf{Multi-modal Instruction Tuning}: We further fine-tune UNIMO-G on millions of pairs of multimodal prompts and images, to improve the capability of faithfully generating images from multimodal inputs.
\end{itemize}

It is worth noting that during both stages of training, only the U-Net component is actively trained, with the MLLM parameters frozen. This strategy ensures that UNIMO-G effectively learns to generate images while retaining the perception knowledge encoded in the pre-trained MLLM.

\subsection{Text-to-Image Pre-training}

\paragraph{Preliminaries} 
We follow the latent diffusion model~\cite{rombach2022high}, utilizing the perceptual compression model (i.e., VQ-VAE) consisting of an image encoder $\mathcal{E}$ and decoder $\mathcal{D}$ to encode the pixel data into the latent space and reverse, such that $\mathcal{D}(\mathcal{E}(x)) \approx x$. The diffusion process is then performed on the latent space, which defines a Markov chain of forward diffusion process $q$ by gradually adding Gaussian noise to the initial latent representations $z_0 = \mathcal{E}(x)$ over $T$ steps. The forward process $q(z_t|z_{t-1})$ at each time step $t$ can be expressed as follows:
\[q(z_t|z_{t-1})=\mathcal{N}(z_t;\sqrt{1-\beta_t}z_{t-1},\beta_{t}I)\]
where $\{\beta_t\}$ is a series of hyper-parameters. 
Diffusion models are trained to learn a conditional U-Net~\cite{ronneberger2015u} denoiser $\epsilon_\theta$ to reverse the diffusion Markov chain, predicting noise with current timestep $t$, noisy latent $z_t$ and generation condition $c$.  The training loss is the mean squared error (MSE) between the predicted noise $\epsilon_\theta(z_t,t,c)$ and the real noise $\epsilon$ in $z_t$:
\[\mathcal{L}=\mathbb{E}_{z_0, c, \epsilon \sim \mathcal{N}(0,1), t}[\| \epsilon - \epsilon_{\theta}(z_t,t,c)] \|^2 \]
Large-scale Chinese text-image pairs are utilized to train the above denoising objective. The condition information $c$ is fed into each cross attention block of the UNet model as:
\[Attn(z_t, c) = softmax(\frac{Q(z_t) \cdot K(c)^T}{\sqrt{d}}) \cdot V(c)\]
where $Q$, $K$ and $V$ denote the query, key and value projections, respectively. $d$ denotes the output dimension of the features. In our model, the condition $c$ is encoded by the pre-trained MLLM.

\paragraph{Pre-training Strategies}
Training a text-to-image diffusion model from scratch presents significant challenges in terms of complexity and resource expenditure. To address these, we introduce an effective training schedule to enhance the efficiency and performance of model training. This schedule encompasses three phases: (1) initial training on a small image corpus to establish foundational visual distribution knowledge; (2) subsequent expansion to large-scale text-image pair training, focusing on text-visual correspondence; (3) culminating in a final phase of training with a small refined corpus, characterized by high visual aesthetics and precise text-image alignment. 
In our experiments, the training of the UNet model is initially conducted using the CC3M dataset~\cite{sharma2018conceptual}. This dataset is chosen for its diverse range of visual concepts coupled with straightforward image descriptions, making it an effective tool for initiating training from scratch. Subsequently, the model undergoes further training with an extensive collection of 300M text-image pairs, aimed at broadening its conceptual understanding and improving its alignment to textual descriptions. The final phase of training involves fine-tuning the model using a meticulously curated corpus, consisting of tens of thousands of high-quality image-text pairs, carefully selected for their superior quality.
Based on the above strategies and our architecture designs, we obtain a powerful Chinese text-to-image generation model, surpassing open-source models like Stable Diffusion and its advanced version SDXL~\cite{podell2023sdxl}. We provide detailed evaluations of our results in Section~\ref{eval_res} and implementation details in Appendix~\ref{detail}.

\subsection{Multimodal Instruction Tuning}
Following the text-to-image pre-training, UNIMO-G is indeed capable of generating images from interleaved images and texts, relying on the perception capabilities of MLLM. However, it is important to note that the pre-training stage primarily focuses on generating images that are semantically consistent with the input representations. As a result, UNIMO-G still face challenges in utilizing the visual features of inputs to faithfully reproduce the contents specified in the image conditions. To address this limitation, we further conduct multimodal instruction tuning to enhance UNIMO-G's ability to faithfully reproduce image contents in diverse contexts.

\begin{figure}
    \centering
    \includegraphics[width=0.95\linewidth]{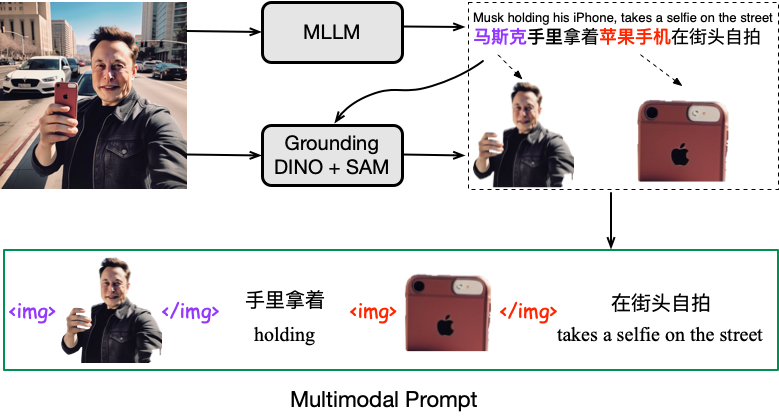}
    \caption{Overview of our data construction pipeline for multi-modal instruction tuning.}
    \label{fig:pipeline}
\end{figure}

\paragraph{Multimodal Prompts} 
To enhance the representativeness of text prompts, we introduce a format for multimodal prompts that are composed of interleaved images and texts. Specifically, entities mentioned in text captions can be substituted with their corresponding images, like ``<img>image of Elon Musk</img> holding his <img>image of iPhone</img>, takes a selfie on the street'', as shown in Figure~\ref{fig:architecture}.  To create pairs of multimodal prompts and images, we have designed a data processing pipeline as illustrated in Figure~\ref{fig:pipeline}. The pipeline first generates captions and extracts entities from the caption by prompting the MLLM. Subsequently, it acquires the corresponding image segment for each entity using a combination of language grounding by Grounding DINO~\cite{liu2023grounding} and image segmentation by SAM~\cite{kirillov2023segment}. Further introduction on the data construction process is provided in Section~\ref{detail}. With a collection of pairs of multimodal prompts and images, UNIMO-G is trained to generate images in accordance with these multimodal instructions. 

\paragraph{Visual-Enhanced Learning} 
In order to better harness the visual features of multi-modal input, we introduce an enhancement to the cross-attention mechanism between the generated objects and the input image entities. This improvement aims to foster a more robust and context-aware understanding of the relationships between generated content and the visual elements within the input images. As stated by Prompt-by-Prompt~\cite{hertz2022prompt}, the cross-attention in text-to-image diffusion models can reflect the positions of each generated object specified by the corresponding text token. Similarly, the visual features of image entities can also be treated as visual tokens. The cross-attention map between the intermediate feature of the noisy latent $z_t$ and the visual token $v$ can be calculated:
\[CA(z_t, v) = Softmax(\frac{Q(z_t) \cdot K(v)^T}{{\sqrt{d}}})\]
where $Q$ and $K$ denote the query and key projections, respectively. For each visual token, we could get an attention map of $h \times w$, where $h$ and $w$ are the spatial dimensions of the latent feature $z_t$. The scores in cross-attention maps represent the amount of information that flows from a visual token to a latent pixel. Therefore, we introduce an additional loss term that encourages the model to ensure that each visual token mainly attends to the image region occupied by the corresponding objects. Specifically, we optimize $z_t$ towards the target that the desired area of the object has large values by penalizing the $L1$ deviation between the attention maps and the corresponding segmentation maps of the entities:
\[\mathcal{L}_{attn} = \frac{1}{N}\sum_{k=1}^{N} \mid CA(z_t, v_k) - M_k \mid\]
where $M_k$ is the segmentation mask of the $k_{th}$ object corresponding to its visual token $v_k$.
Through this training process, UNIMO-G learns to effectively harness the visual features of input images to faithfully reproduce the corresponding content. 

\section{Related Work}
\paragraph{Text-to-Image Diffusion Generation}
The incorporation of diffusion models into text-to-image synthesis represents a notable advancement in computational creativity~\cite{ho2020denoising, song2020denoising,li2022upainting}. Models like GLIDE~\cite{nichol2021glide} and DALL-E 2~\cite{ramesh2022hierarchical}, which utilize CLIP image embeddings, have substantially progressed in producing images that are both diverse and semantically coherent with textual inputs. Imagen~\cite{saharia2022photorealistic} underscores the importance of language comprehension, proposing the integration of a large T5 language model to enhance semantic representation. The Latent Diffusion Model~\cite{rombach2022high} addresses computational constraints by generating images from text-conditioned, low-dimensional latent spaces. Our proposed framework builds upon the principles of the Latent Diffusion Model, leveraging its computational efficiency and scalability.

\paragraph{Subject-Driven Image Generation}
Following the success of generating high quality images from text descriptions, recent studies have explored subject-driven generation techniques. Models like DreamBooth~\cite{ruiz2023dreambooth}, textual-inversion~\cite{gal2022image}, and custom-diffusion~\cite{kumari2023multi} use optimization-based methods to embed subjects into diffusion models. This is achieved by either fine-tuning the model weights or inverting the subject image into a text token that encodes the subject identity. Some works have explored tuning-free methods. ELITE~\cite{wei2023elite} and InstantBooth~\cite{shi2023instantbooth} project reference images into word embeddings and inject reference image patch features into cross-attention layers to enhance local details. PhotoMaker~\cite{li2023photomaker} focuses on the generation of human portraits by extracting a stacked ID embedding from multiple ID images.
Despite impressive results for single-object customization, their architecture design restricts their scalability to multiple subject settings. 
Models like Subject-Diffusion~\cite{ma2023subject} and FastComposer~\cite{xiao2023fastcomposer} are designed for multi-entity subject-driven generation. They enhance text conditioning within diffusion models by incorporating subject embeddings extracted from separate image encoders.
Yet, a prevalent limitation of these approaches is their inclination to separate textual and visual guidance, thereby constraining the efficacy of joint modality integration. 

\paragraph{Generating with Multi-modal Language Models}
Multimodal Large Language Models (MLLMs) have significantly broadened the capabilities of language models to process various modalities~\cite{liu2023llava,li-etal-2021-unimo,li-etal-2022-unimo,wang2023cogvlm,driess2023palm}. These models inherently facilitate interleaved vision-language input, effectively handling multiple images.  
Models such as GILL~\cite{koh2023generating}, Emu~\cite{sun2023generative}, and DreamLLM~\cite{dong2023dreamllm} specialize in interleaved vision-language generation by aligning the output space of MLLMs with the diffusion image decoder. However, these methods primarily align at a semantic level and may struggle with detailed, subject-driven image generation. 
BLIP-Diffusion~\cite{li2023blip} synthesizes images by composing subjects with random backgrounds, endowing it with zero-shot, subject-driven text-to-image generation capabilities. However, its specific input template and training process limit scalability for multiple entities.
KOSMOS-G~\cite{pan2023kosmos}, a model closely related to our work, leverages a MLLM to encode interleaved text-visual inputs, and the U-Net of Stable Diffusion (SD) v1.5 as the image decoder. The key component of KOSMOS-G is an AlignerNet, which is trained solely on textual data to align the output embedding space of the frozen SDv1.5 U-Net with the MLLM.
In contrast, our approach centers on training the U-Net model end-to-end specifically for multimodal diffusion, significantly enhancing both the faithfulness and relevance of generated images in multimodal contexts. 
Differing from alignment-based approaches, our two-stage training strategy markedly improves the model's proficiency in following multimodal instructions, particularly in complex multi-entity scenarios. 

\section{Experiments}
\label{exp}
In this section, we first introduce the implementation details and settings of experiments. Then we present the evaluation results in both text-driven and subject-driven scenarios. Last, we further analyze the results with quantitative ablation studies.

\subsection{Implementation Details}
UNIMO-G is composed of a 7.8B-parameter MLLM encoder, following the MiniGPT-4 architecture~\cite{zhu2023minigpt}, and a 4B-parameter denoising U-Net, totaling approximately 11.8B parameters. The MLLM is pretrained on a large-scale Chinese multimodal corpus, comprising text, image-caption pairs, and interleaved image-text data. The U-Net architecture includes 5 downsampling and 5 upsampling blocks with channel sizes [640, 1280, 1280, 2560, 2560], and a cross-attention mechanism with 4096 dimensions and 16 heads. The image auto-encoder, based on the LDM framework, has been optimized for our specific image corpus. The detail training process and data construction are further introduced in Appendix~\ref{detail}.

\subsection{Evaluation Results}
\label{eval_res}
UNIMO-G demonstrates a unified image generation capability for both text-driven and subject-driven image generation, as shown in Figure~\ref{fig:showcase}. In the following, we will evaluate the performance of UNIMO-G from different aspects.

\paragraph{Text-to-Image Generation}
For text-to-image generation, we used 30,000 captions randomly sampled from the MS-COCO~\cite{lin2014microsoft} validation set, translating them into Chinese to align with UNIMO-G's input requirements. Images were generated at 512x512 pixels and resized to 256x256 for evaluation using the FID-30k metric, a standard in the field. Our model employs a classifier-free guidance scale of 5.0 and 50 DDIM inference steps for diffusion sampling. As shown in Table~\ref{tab:main_exp}, UNIMO-G greatly surpasses other Vision-Language to Image (VL2I) models in performance.

\begin{figure}[t]
    \centering
    \includegraphics[width=0.8\linewidth]{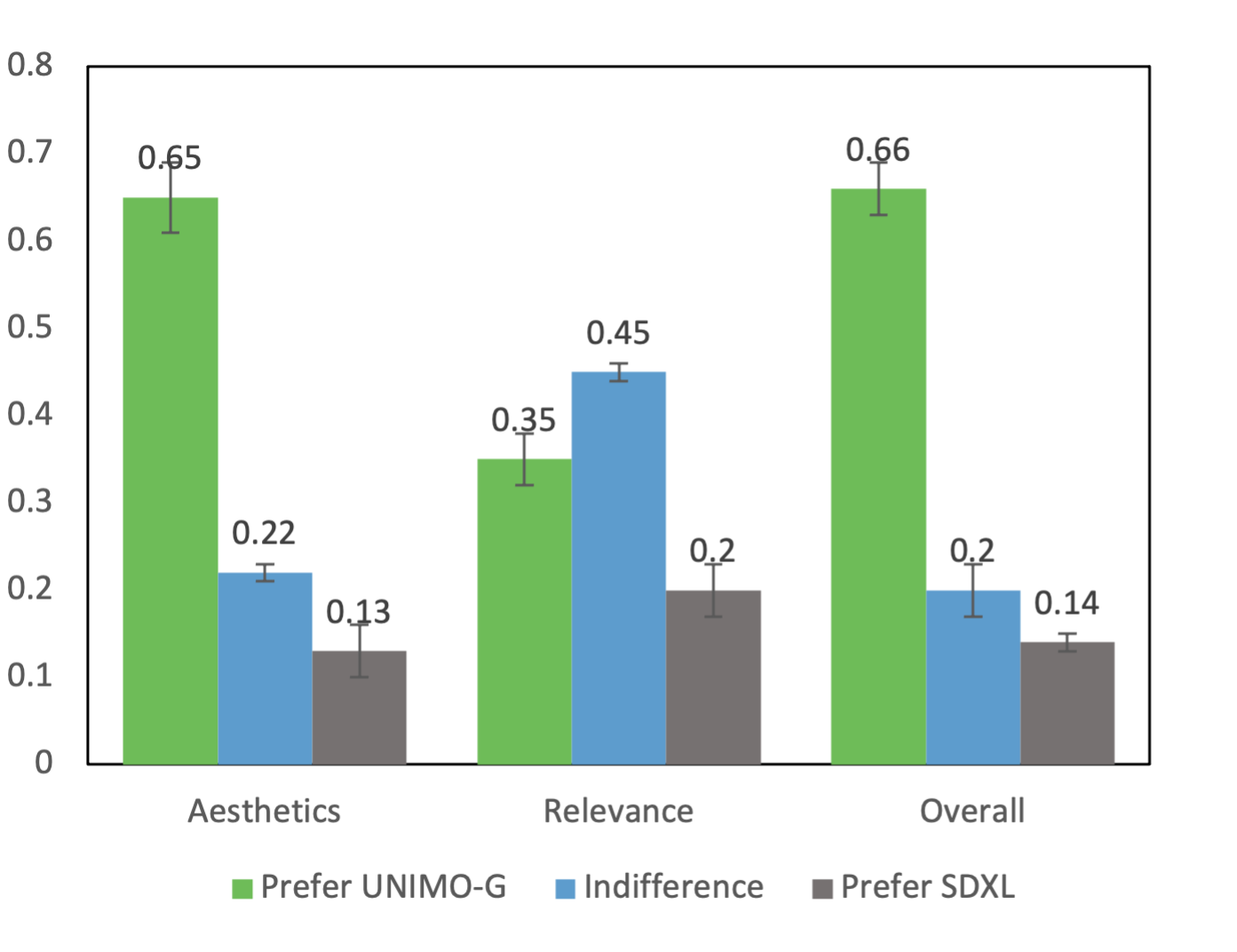}
    \caption{Comparison of UNIMO-G and SDXL by human evaluation. The mean and standard deviation are shown in the figure.}
    \label{fig:t2i-comparison}
\end{figure}

To further validate our model, we conducted a human evaluation comparing UNIMO-G with SDXL~\cite{podell2023sdxl}, a leading open-source model. We established a comprehensive bilingual benchmark, encompassing 312 prompts (162 from DrawBench and 150 user queries randomly sampled from the online platform\footnote{https://yige.baidu.com/}). The DrawBench prompts were filtered to exclude language-specific ones. All prompts are manually translated and carefully proofread to achieve the final parallel Chinese and English set. Three independent evaluators rated the images from UNIMO-G and SDXL by selecting the model they prefer, focusing on aspects of image aesthetics, image-text relevance, and overall quality, respectively. The results in Figure~\ref{fig:t2i-comparison} demonstrate UNIMO-G's substantial superiority in all aspects. Some examples are shown in Figure~\ref{fig:t2i-cases}.

\begin{table}[!t]
\centering
\small
\begin{tabular}{lll}
\toprule
\textbf{Methods} & \textbf{} & \textbf{FID} \\ \midrule
\multicolumn{3}{c}{\textit{T2I Models}}     \\ \midrule
GLIDE~\cite{nichol2021glide}            &           & 12.24        \\ 
DALL-E 2~\cite{ramesh2022hierarchical}         &           & 10.39        \\ 
SDv1.5~\cite{rombach2022high}          &           & 9.34         \\ 
Imagen~\cite{saharia2022photorealistic}           &           & 7.27         \\ 
SDXL~\cite{podell2023sdxl}    &         &   11.93   \\   \midrule
\multicolumn{3}{c}{\textit{VL2I Models}}   \\  \midrule
GILL~\cite{koh2023generating}             &           & 12.20        \\ 
Emu~\cite{sun2023generative}             &           & 11.66        \\ 
KOSMOS-G~\cite{pan2023kosmos}         &           & 10.99        \\
UNIMO-G          &           & 8.36         \\ \bottomrule
\end{tabular}
\caption{Zero-shot FID-30K comparisons on MS-COCO 256x256.}
 \label{tab:main_exp}
\end{table}

\paragraph{Single-Entity Subject-Driven Generation}
For single-entity subject driven generation, we evaluate UNIMO-G on DreamBench~\cite{ruiz2023dreambooth}. DreamBench comprises 30 subjects with 25 prompt templates, yielding 750 unique prompts that test skills such as re-contextualization, modification, and accessorization. We follow prior work to generate four images for each prompt, creating a total of 3,000 images for a comprehensive assessment. We employed DINO and CLIP-I metrics for subject fidelity evaluation and CLIP-T for image-text relevance assessment. A classifier-free guidance scale of 5.0 and 50 DDIM inference steps were used for sampling. UNIMO-G, accepting a single image input, utilized the same images as KOSMOS-G~\cite{pan2023kosmos} for a consistent comparison. As indicated in Table~\ref{tab:single}, UNIMO-G in a zero-shot setting surpasses other models like Textual Inversion~\cite{gal2022image}, DreamBooth~\cite{ruiz2023dreambooth}, BLIP-Diffusion~\cite{li2023blip}, and Re-Imagen~\cite{chen2022re}, and marginally outperforms KOSMOS-G. Notably, our model demonstrates a significant improvement in balancing image-text relevance and image fidelity compared to the closely related KOSMOS-G. We observed that existing methods tend to prioritize image information over textual input. This tendency occasionally leads to a diminished focus on semantic content, favoring subject reconstruction. Thanks to our two-stage end-to-end learning framework, UNIMO-G maintained high image fidelity and achieved the highest CLIP-T score for image-text relevance, indicating a strong capability in following multi-modal instructions.

\begin{figure*}[!ht]
    \centering
    \includegraphics[width=0.9\linewidth]{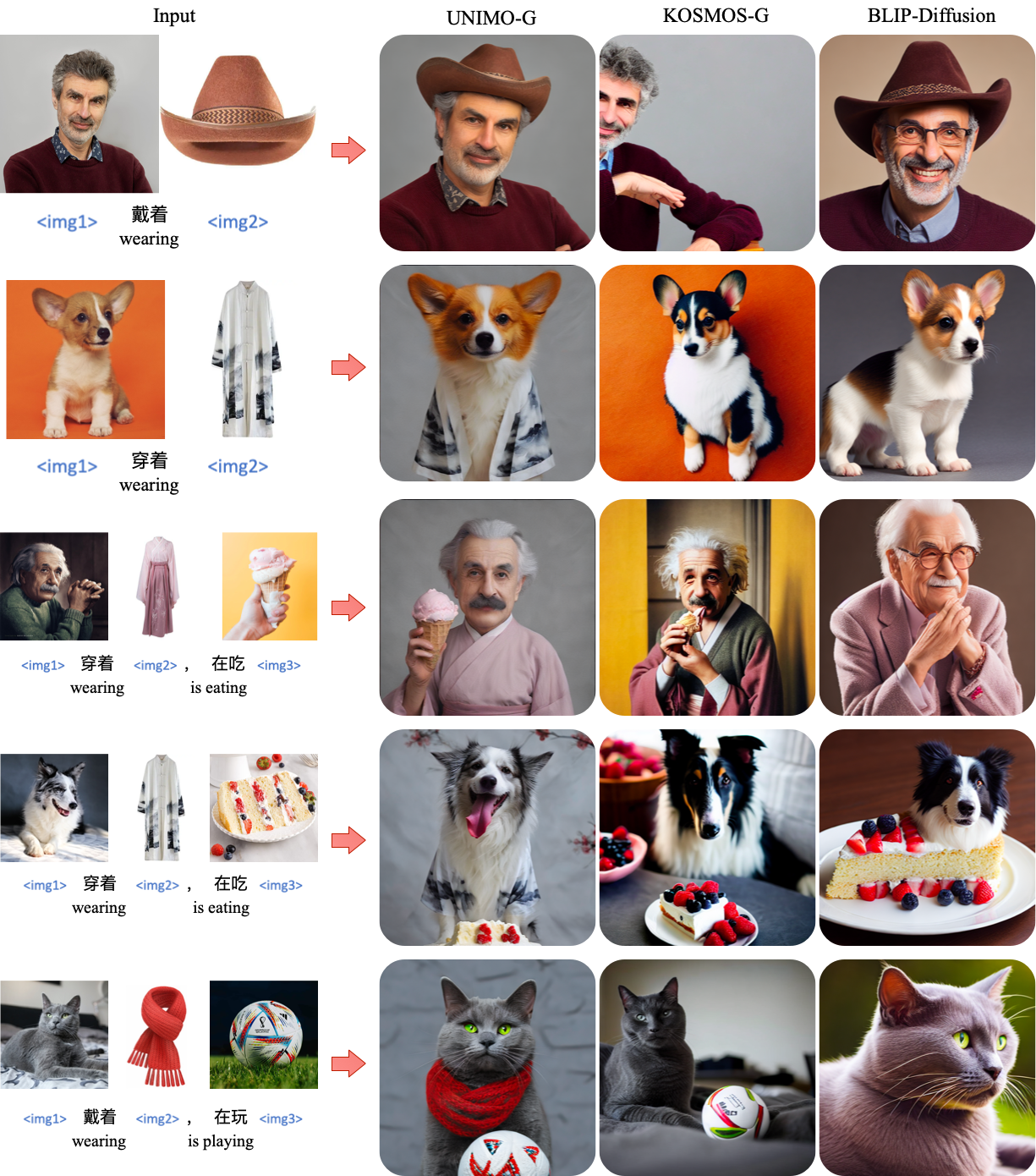}
    \caption{Comparison with baselines for multi-entity subject-driven image generation.}
    \label{fig:comparsion}
\end{figure*}

\paragraph{Multi-Entity Subject-Driven Generation}
UNIMO-G exhibits exceptional performance in zero-shot multi-entity subject-driven generation. To evaluate this capability, we established \textit{MultiBench}, a novel benchmark specifically designed for multi-entity subject-driven generation assessment. MultiBench includes four object categories: living objects (humans and animals), food, wearable items, and toys, each containing 10 different objects. We developed five prompt templates for composing scenarios with 2 and 3 objects, resulting in a total of 2,300 distinct prompts. Details are introduced in the Appendix~\ref{sec:multibench}.
For each prompt, four images were generated, culminating in 9,200 images for an exhaustive evaluation. We conducted image similarity analysis using DINO and CLIP-I metrics, alongside text relevance assessments using CLIP-T. Image similarity was determined by averaging the similarities between the generated image and each of the two or three subjects. The results, as shown in Table~\ref{tab:multi}, indicate that UNIMO-G outperforms BLIP-Diffusion and KOSMOS-G in terms of both image similarity and textual relevance. Some comparison examples are shown in Figure~\ref{fig:comparsion}. This demonstrates UNIMO-G's superior capability to accurately capture subject information from input images and effectively follow multi-modal instructions. More examples are shown in Figures~\ref{fig:selected-multi-entity-cases} and~\ref{fig:selected-multi-entity-cases2}.

\begin{table}[]
\centering
\small
\begin{tabular}{@{}lllll@{}}
\toprule
\textbf{Methods}  & \textbf{DINO} & \textbf{CLIP-I} & \textbf{CLIP-T} & \textbf{Avg} \\ \toprule
\multicolumn{5}{c}{\textit{Fine-Tuning Methods}}                                              \\ \midrule
Textual Inversion & 0.569         & 0.780           & 0.255           & 0.535         \\ 
DreamBooth        & 0.668         & 0.803           & 0.305           & 0.592         \\ 
BLIP-Diffusion    & 0.670         & 0.805           & 0.302           & 0.592         \\ \midrule
\multicolumn{5}{c}{\textit{Zero-Shot Methods}}                                                \\ \midrule
Re-Imagen         & 0.600         & 0.740           & 0.270           & 0.537         \\ 
BLIP-Diffusion    & 0.594         & 0.779           & 0.300           & 0.558         \\ 
KOSMOS-G          & \textbf{0.694}         & \textbf{0.847}           & 0.287           & 0.609         \\ 
UNIMO-G           & 0.668         & 0.841           & \textbf{0.329}         & \textbf{0.613}       \\ 
\ \ w/o Tuning & 0.371             & 0.717                & 0.306                & 0.465              \\
\ \ w/o VisualEnh & 0.617             & 0.815                & 0.329                & 0.587              \\ \bottomrule
\end{tabular}
\caption{Comparisons of single-entity subject-driven image generation on DreamBench. \textit{Avg} denotes the average scores of DINO, CLIP-I and CLIP-T.}
\label{tab:single}
\end{table}

\begin{table}[]
\centering
\small
\begin{tabular}{@{}lllll@{}}
\toprule
\textbf{Methods}  & \textbf{DINO} & \textbf{CLIP-I} & \textbf{CLIP-T} & \textbf{Avg} \\ \toprule
BLIP-Diffusion & 0.410  & 0.648  & 0.249  & 0.436 \\
KOSMOS-G          & 0.419  & \textbf{0.671}  & 0.283  & 0.458 \\
UNIMO-G    & \textbf{0.436}  & 0.665  & \textbf{0.298}   & \textbf{0.466}      \\
\ \ w/o Tuning & 0.235             & 0.583                & 0.240                & 0.353            \\
\ \ w/o VisualEnh & 0.399             & 0.631                & 0.276               & 0.435              \\ 
\bottomrule
\end{tabular}
\caption{Comparisons of multi-entity subject-driven image generation on MultiBench.}
\label{tab:multi}
\end{table}

\begin{figure}[t]
    \centering
    \includegraphics[width=0.8\linewidth]{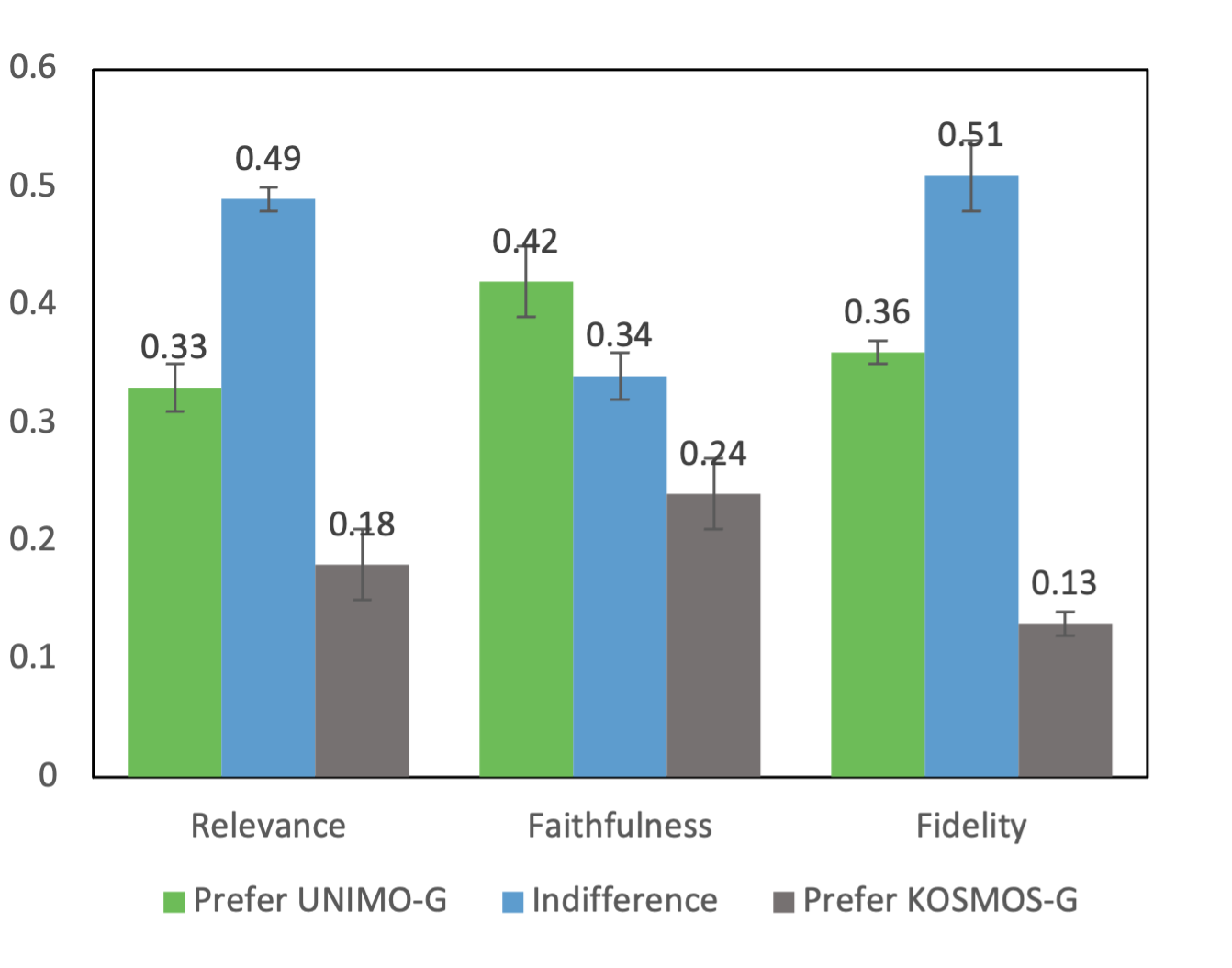}
    \caption{Comparison of UNIMO-G and KOSMOS-G on MultiBench by human evaluation. The mean and standard deviation are shown in the figure.}
    \label{multibench_human_eval}
\end{figure}

To further validate our model, we conducted a human evaluation comparing UNIMO-G with KSOMOS-G by sampling 200 prompts from MultiBench. Three raters are presented with two sets of images generated by UNIMO-G and the compared model. They are asked to compare these images from three dimensions of semantic relevance, visual faithfulness and image fidelity, and then select the model they prefer, or indifference. Throughout the process, raters are unaware of which model the image is generated from. The results in Figure~\ref{multibench_human_eval} show that human raters greatly prefer UNIMO-G over KOSMOS-G on all aspects, which further validate the effectiveness of our approach in generating high-quality, personalized images from free-form multimodal prompts.

\subsection{Analysis}

\paragraph{Effectiveness of Multi-modal Instruction Tuning}
UNIMO-G, prior to multi-modal instruction tuning (denoted as ``w/o Tuning''), also demonstrates the capability to generate images from multi-modal prompts based on the MLLM. Nonetheless, it initially falls short in accurately reproducing input images. To evaluate the effectiveness of multi-modal instruction tuning, we compared the performance of UNIMO-G with and without this tuning in single-entity and multi-entity subject-driven generation tasks. The comparison results in Table~\ref{tab:single} and Table~\ref{tab:multi} reveal that multi-modal instruction tuning substantially enhances image similarity metrics (DINO and CLIP-I) in both single and multi-entity scenarios. This improvement indicates that after tuning, UNIMO-G more effectively leverages visual features from the inputs, thereby accurately replicating the content defined in the image conditions. Furthermore, UNIMO-G with instruction tuning also shows obvious advancements in textual relevance, as evaluated by the CLIP-T metric. This indicates that the tuning process not only bolsters visual faithfulness but also amplifies the model's ability to follow multimodal instructions.

\paragraph{Effectiveness of Visual Enhancement Learning}
The incorporation of a visual-enhanced learning strategy during multimodal instructional tuning significantly improves the visual alignment between input and output images. To quantify this effect, we conducted an ablation study by omitting the visual enhancement component during multimodal tuning (denoted as ``w/o VisualEnh'') and assessed its impact on both single-entity and multi-entity generation tasks. The results, as detailed in Tables~\ref{tab:single} and~\ref{tab:multi}, demonstrate that the visual-enhanced learning strategy markedly boosts the performance in image similarity metrics (DINO and CLIP-I), across both single and multi-entity scenarios. Notably, it also improves image-text alignment in multi-entity scenarios by reducing entity blending or missing.

\section{Conclusion}
This paper presents UNIMO-G, a simple multimodal conditional diffusion framework designed to process multimodal prompts that interleave text and visual inputs. It demonstrates exceptional proficiency in text-to-image generation and zero-shot subject-driven synthesis, and is particularly adept at producing high-fidelity images from intricate multi-modal prompts with multiple image entities. In comparison to standard text-conditional diffusion models, UNIMO-G significantly enhances visual controllability in image generation. Thanks to our two-stage training strategy, UNIMO-G also outperforms existing VL-to-image models, especially on the ability to follow complex multimodal instructions. Overall, UNIMO-G showcases the potential for more nuanced and controlled image generation processes.

\section{Limitations}
Our model suffers from some common failures of text-driven and subject-driven generation models, such as inaccuracies in context synthesis, difficulties in complex composition, and a shortfall in visual faithfulness, particularly in multi-entity image generation tasks. Additionally, there exists an inherent risk associated with the misuse of such technology, notably in the creation of deepfakes, which raises ethical concerns.
Despite the limitations and risks, the proposed framework still demonstrates considerable promise in facilitating more nuanced and controlled processes in image generation.

\bibliography{custom}
\bibliographystyle{acl_natbib}

\appendix

\section{Multi-modal Large Language Model}
\label{sec:mllm}
\begin{figure}[!ht]
    \centering
    \includegraphics[width=0.95\linewidth]{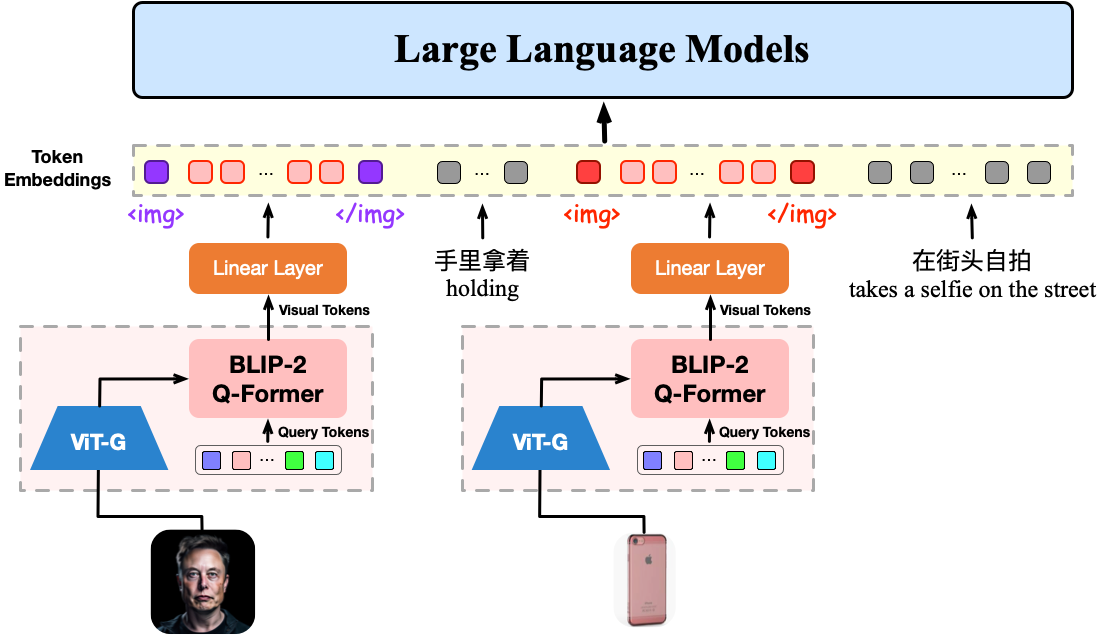}
    \caption{Illustration of the encoding of multimodal prompts by the MLLM.}
    \label{fig:mllm}
\end{figure}

The structure of our MLLM is similar to MiniGPT-4~\cite{zhu2023minigpt} and BLIP-2~\cite{li2023blip}, which consists of a vision encoder with a pretrained ViT-G and BLIP-2 Q-Former, a single linear projection layer, and a pre-trained 6B Transformer-based LLM. The MLLM totally contains about 7.8B parameters. The encoding process of multimodal prompts is illustrated as in Figure~\ref{fig:mllm}. Each image is firstly encoded as 32 visual tokens by the visual encoder, and then linearly transformed into the LLM embedding space. The pre-trained LLM is utilized as the context encoder and decoder. Our MLLM underwent training on web-scale multimodal corpora, comprising text corpora, image-caption pairs, and interleaved data of images and texts.

\section{Implementation Details}
\label{detail}
The training process include text-to-image pre-training and multi-modal instruction tuning.

\paragraph{Text-to-Image Pre-training}
Our model's pre-training involves three stages, each utilizing distinct datasets:
\begin{enumerate}
    \item Initial Training with CC3M Dataset: The CC3M dataset, consists of about 3.3M image-description pairs, was translated into Chinese using the Baidu Translation API. The model is trained from scratch at 256x256 using the AdamW optimizer with a weight decay of 0.01, a learning rate of 5e-5, and a batch size of 40x256 for 100K steps, which costs about 300 A100 GPU days.
    \item Expansion with Large-Scale Chinese Data: We incorporate about 300M Chinese text-image pairs from multiple datasets, including LAION-2B~\cite{schuhmann2022laion}, COYO-700M~\cite{kakaobrain2022coyo-700m}, Conceptual Captions~\cite{changpinyo2021conceptual} and a series of internal Chinese datasets. The English captions are translated into Chinese. This stage, with a constant learning rate of 5e-5, initially training at 256x256 resolution for 500K steps with a batch size of 40x256, then progresses to 512x512 for 200K steps with a batch size of 12x256, which costs about 3000 GPU days.
    \item Refinement with High-Quality Corpus: The final stage focuses on fine-tuning on high-quality corpus, and continues training at 512x512 and 1024x1024 resolutions. Selection criteria include an aesthetic threshold above 6.5 (LAION-Aesthetics V2) and an image-text similarity score over 0.5 (CLIP-base-32), resulting in about 1M pairs. A multi-scale training strategy dividing the image size into 5 buckets [0.5, 0.75, 1.0, 1.5, 2.0] supports various image aspect ratios. The final stage uses a learning rate of 1e-5 for 200K steps with a batch size of 3072, which costs about 1000 GPU days.
\end{enumerate}

\paragraph{Multi-modal Instruction Tuning}
We developed a multimodal-to-image instruction tuning dataset utilizing the 1M high-quality image-text pairs. The process, depicted in Figure~\ref{fig:pipeline}, involves: (1) Generation of captions and extraction of entity tokens from captions by the MLLM; (2) Identifying entity detection boxes via Grounding DINO~\cite{liu2023grounding}; (3) Segmentation and extraction of regions corresponding to each entity by SAM~\cite{kirillov2023segment}; (4) Randomly substitution of entity tokens with their corresponding image segments. This process effectively transforms the original text-image pairs into multimodal-image pairs, which are then employed for refining multimodal instruction tuning. We maintain a learning rate of 1e-5, training for 200K steps with a batch size of 3072. To preserve text-to-image generation capabilities, 10\% of the training uses original textual captions. This process totally costs about 800 A100 GPU days.

\begin{table*}[!ht]
\begin{minipage}[l]{0.3\textwidth}
\centering
\small
\begin{tabular}{ll}
\toprule
Number of Entities  &  Ratios \\
\midrule
0 & 16\% \\
1 & 36.8\% \\
2 & 25.8\% \\
3 & 12.3\% \\
>=4 & 9.1\% \\
\bottomrule
\end{tabular}
\captionof{table}{Statistics of the number of subject entities in our constructed multi-modal prompts.}
\label{tab:multimodal_prompts}
\end{minipage}
\hspace{10mm}
\begin{minipage}[r]{0.6\textwidth}
\centering
\small
\begin{tabular}{@{}llllll@{}}
\toprule
\textbf{Num of Entities} & \textbf{Methods}  & \textbf{DINO} & \textbf{CLIP-I} & \textbf{CLIP-T} & \textbf{Avg} \\
\multirow{3}{*}{2} & BLIP-Diffusion & 0.455  & 0.675  & 0.249  & 0.460 \\
& KOSMOS-G          & 0.465  & \textbf{0.704}  & 0.279  & 0.483 \\
& UNIMO-G    & \textbf{0.485}  & 0.699  & \textbf{0.293}   & \textbf{0.492}      \\
\midrule
\multirow{3}{*}{3} & BLIP-Diffusion & 0.344  & 0.608  & 0.249  & 0.400 \\
& KOSMOS-G          & 0.350  & \textbf{0.621}  & 0.287  & 0.419 \\
& UNIMO-G    & \textbf{0.364}  & 0.614  & \textbf{0.306}   & \textbf{0.428}      \\
\bottomrule
\end{tabular}
\captionof{table}{Comparisons of the performance with different number of subject entities in a multi-modal prompt on MultiBench.}
\label{tab:compare_entity}
\end{minipage}
\end{table*}

\begin{figure}[!ht]
    \centering
    \includegraphics[width=0.95\linewidth]{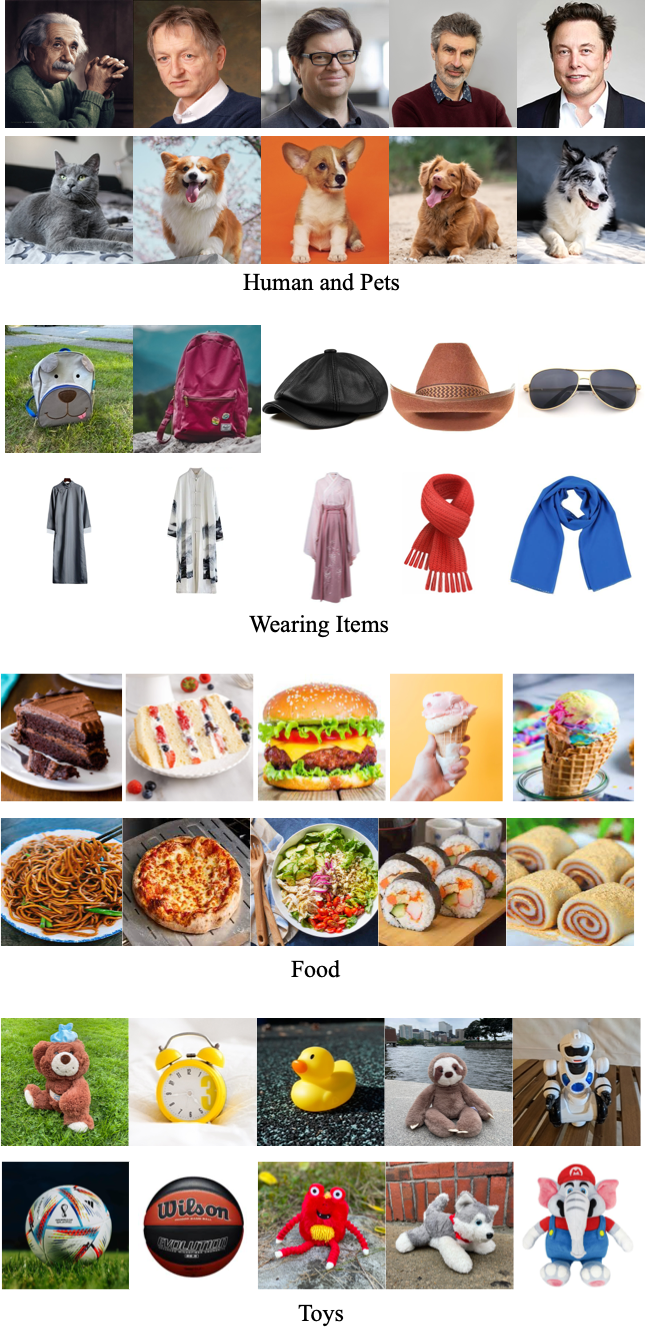}
    \caption{Illustration of images in MultiBench.}
    \label{fig:multibench}
\end{figure}

\section{Analysis of Multi-modal Prompts}
In our analysis of multi-modal prompts, we meticulously examined the composition of our dataset, finding an average of $42.48$ text tokens and $1.67$ subject entities per prompt. The distribution of entity counts revealed a predominance of images featuring fewer than three subject entities, with a minor proportion exceeding this count. The ratios of different number of entities are shown as in Table~\ref{tab:multimodal_prompts}.

In our experiments, we have assessed the model's performance using DreamBench for single-entity prompts and MultiBench for multi-entity scenarios, specifically focusing on two and three entities. We separately evaluate the performance on two and three entities as shown in Table~\ref{tab:compare_entity}. The results, as detailed, showcase that our model, UNIMO-G, consistently surpasses baseline models in handling both two and three-entity multi-modal prompts. This clear performance advantage underscores the effectiveness of our approach across varying levels of complexity.

\begin{table}[t]
\centering
\small
\begin{tabular}{l}
\toprule
a \{living object\} wearing \{wearing\} \\
a \{living object\} is playing with \{toy\} \\
a \{living object\} is eating \{food\} \\
a \{living object\} wearing \{wearing\}, is playing with \{toy\} \\
a \{living object\} wearing \{wearing\}, is eating \{food\} \\
\bottomrule
\end{tabular}
\caption{Templates for multi-entity subject-driven generation in MultiBench.}
\label{tab:multibench}
\end{table}

\section{Images and Prompts of MultiBench}
\label{sec:multibench}
\begin{figure*}
    \centering
    \includegraphics[width=0.95\linewidth]{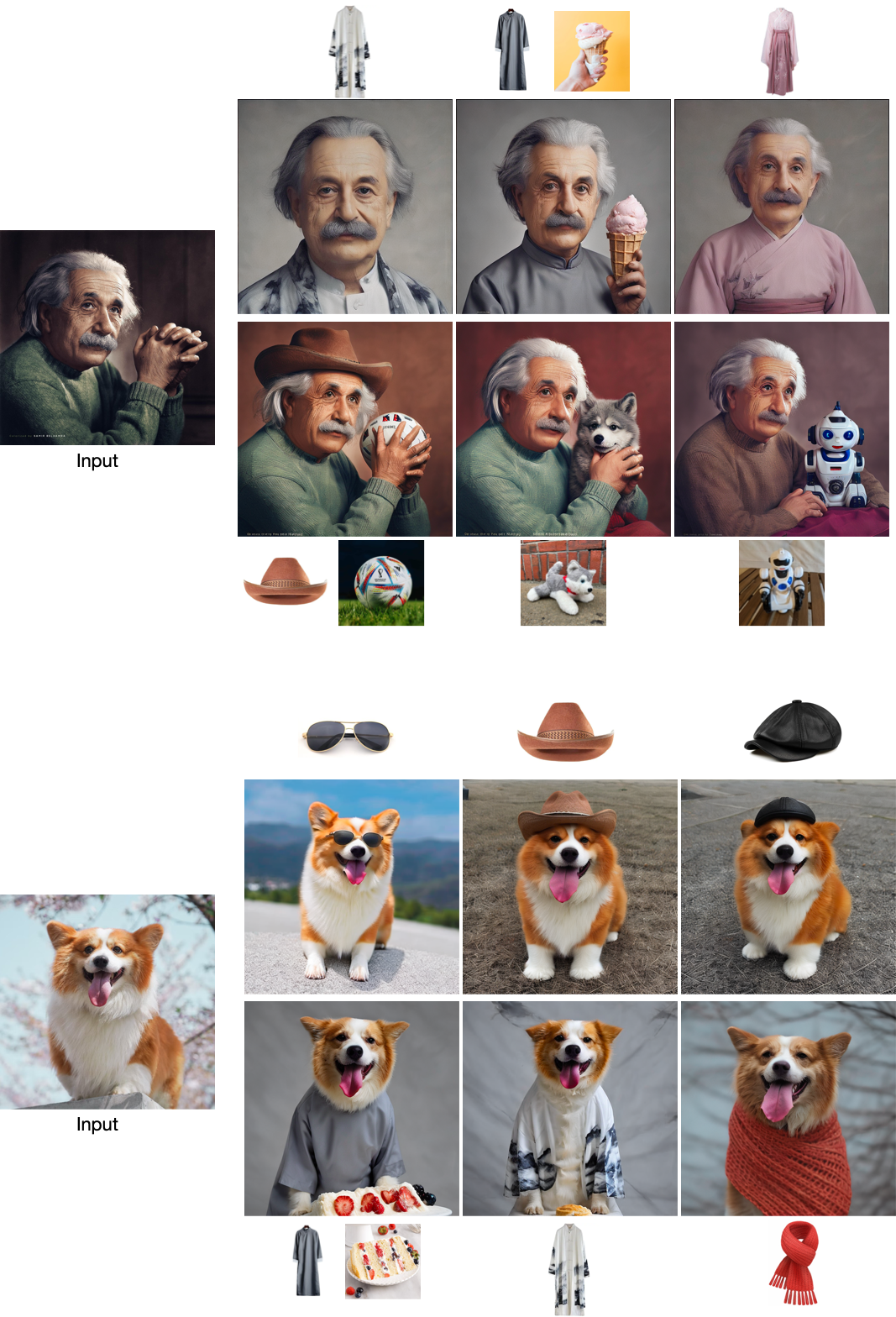}
    \caption{Examples of multi-entity subject-driven image generation on MultiBench by UNIMO-G.}
    \label{fig:selected-multi-entity-cases}
\end{figure*}

\begin{figure*}
    \centering
    \includegraphics[width=0.95\linewidth]{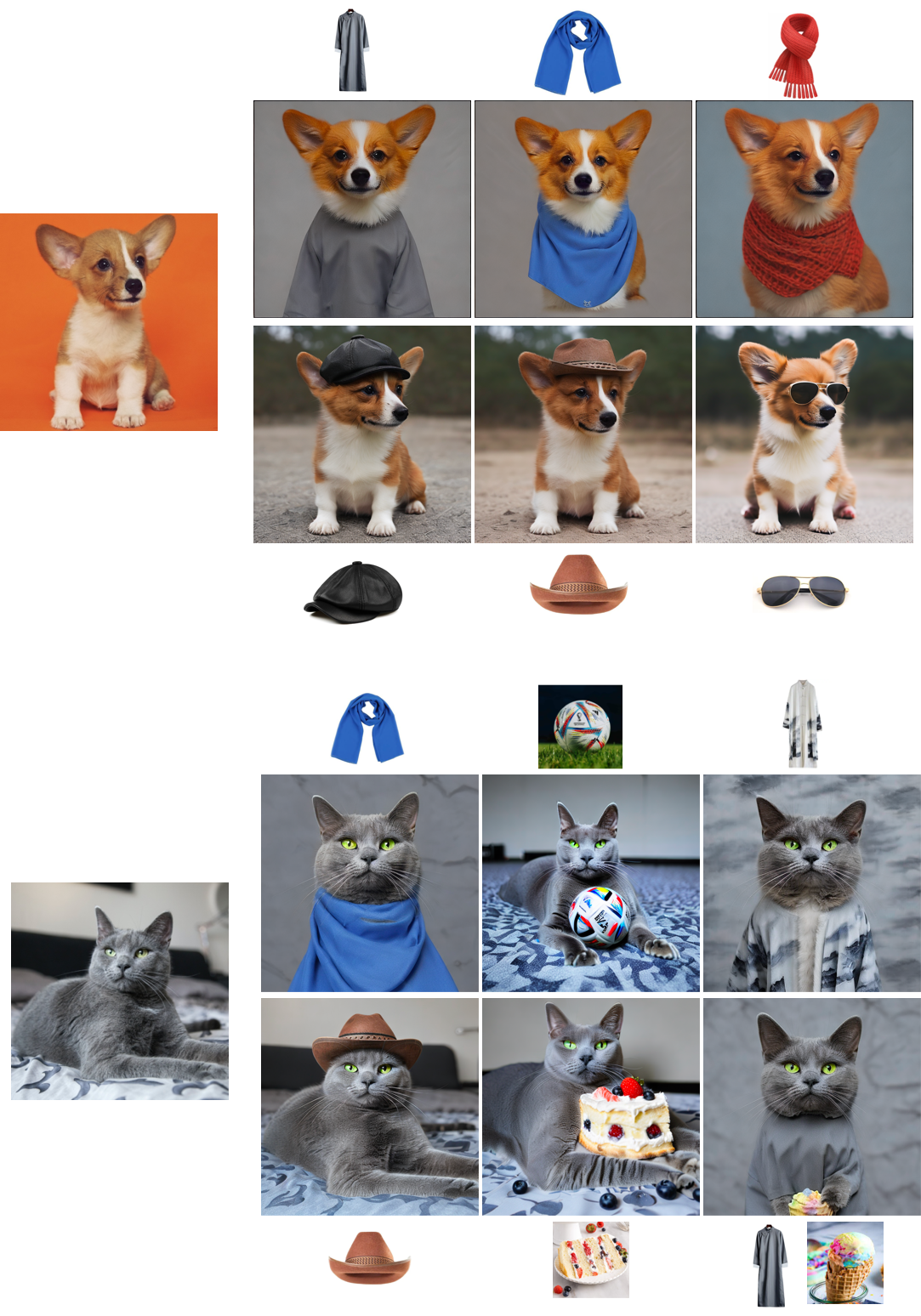}
    \caption{Examples of multi-entity subject-driven image generation on MultiBench by UNIMO-G.}
    \label{fig:selected-multi-entity-cases2}
\end{figure*}

MultiBench contains 4 categories of objects, each including 10 objects. The images in MultiBench are shown in Figure~\ref{fig:multibench}. Five temples are designed to compose multi-modal prompts, as shown in Table~\ref{tab:multibench}. Some examples generated by UNIMO-G on MultiBench are shown in Figure~\ref{fig:selected-multi-entity-cases} and Figure~\ref{fig:selected-multi-entity-cases2}.

\begin{figure*}
    \centering
    \includegraphics[width=0.95\linewidth]{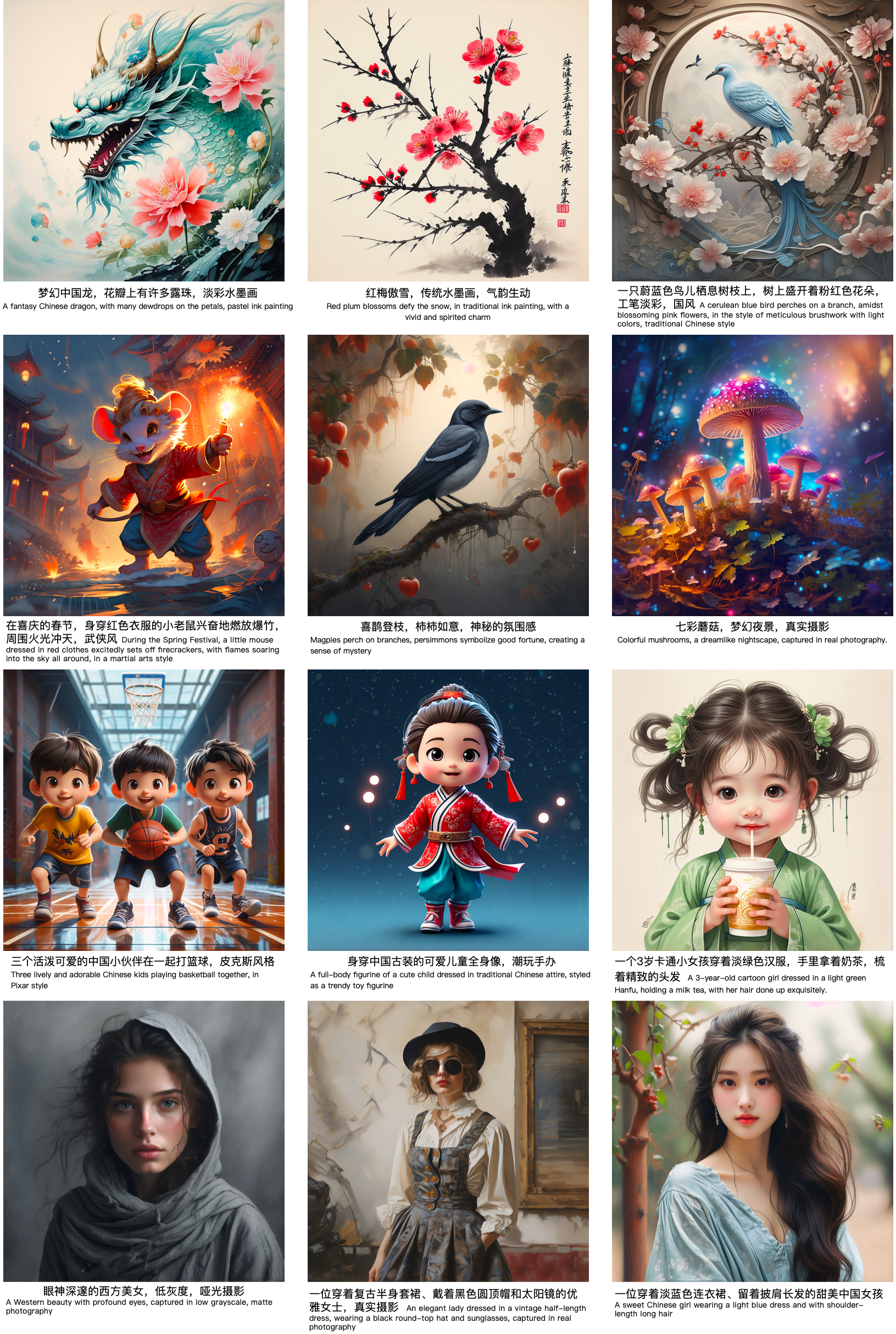}
    \caption{Examples of text-to-image generation by UNIMO-G.}
    \label{fig:t2i-cases}
\end{figure*}

\end{document}